# Kinship in Speech: Leveraging Linguistic Relatedness for Zero-Shot TTS in Indian Languages

*Utkarsh Pathak*[1,*], *Chandra SaiKrishna Gunda*[1,*], *Anusha Prakash*[2], *Keshav Agarwal*[1], *Hema A. Murthy*[1]

[1]Indian Institute of Technology, Madras, India
[2]Independent Researcher, India

`utkarshpathak16@gmail.com, saikrishnag432@gmail.com, anushaprakash90@gmail.com, kshvgrwl@gmail.com,`
`hema@cse.iitm.ac.in`

## Abstract

Text-to-speech (TTS) systems typically require high-quality studio data and accurate transcriptions for training. India has 1369 languages, with 22 official using 13 scripts. Training a TTS system for all these languages, most of which have no digital resources, seems a Herculean task. Our work focuses on zero-shot synthesis, particularly for languages whose scripts and phonotactics come from different families. The novelty of our work is in the augmentation of a shared phone representation and modifying the text parsing rules to match the phonotactics of the target language, thus reducing the synthesiser overhead and enabling rapid adaptation. Intelligible and natural speech was generated for Sanskrit, Maharashtrian and Canara Konkani, Maithili and Kurukh by leveraging linguistic connections across languages with suitable synthesisers. Evaluations confirm the effectiveness of this approach, highlighting its potential to expand speech technology access for under-represented languages.

**Index Terms**: zero-shot synthesis, unseen Indian languages, common label set (CLS), low resource, unified parser.

## 1. Introduction

The Indian subcontinent is home to diverse cultures and languages, reflecting its rich linguistic heritage. Unlike the European Union, which primarily uses 3 to 4 major scripts, India has around 66 scripts, of which 13 are used in the 22 official languages. Furthermore, 121 Indian languages are spoken by at least 10,000 individuals [1]. India's linguistic landscape, shaped by geographical, social, cultural, and migration factors, encompasses languages from Indo-Aryan (IA), Dravidian (DR), Sino-Tibetan (ST), and Austroasiatic families, alongside 19,569 dialects [1]. This diversity, while culturally enriching, creates challenges in communication, education, healthcare, and governance.

There has been significant progress made in the development of text-to-speech (TTS) models, leading to substantial improvements in speech synthesis quality [2–8]. Despite these advancements, building TTS for low-resource languages is still challenging due to the scarcity of high-quality TTS data. Recent progress in multilingual synthesis [9–12] has introduced various frameworks designed to generate zero-shot multilingual, multi-speaker speech. However, these approaches predominantly rely on large-scale datasets, and comprehensive evaluations on individual languages remain limited. This limitation is particularly evident in the Indian context, as even most official languages are under-represented in these large-scale multilingual models.

For low resource scenarios (common in most NLP tasks), most systems use resource-rich languages to synthesise the target language or use cross-lingual transfer learning [13, 14]. Another widely adopted strategy is joint training across multiple languages, enabling mutual support through shared linguistic features [15]. Also, a long-standing goal that has gained renewed interest recently is the development of language-independent models, which would take into account all the typological variations [12, 16].

Recent efforts on multilingual Indic TTS systems have focused on establishing a unified linguistic front-end [15, 17–19]. TTS models have been trained by grouping closely related languages based on phylogenetic or areal studies [15] and language families [17, 19]. Zero or minimal adaptation data is then used to synthesise speech in the target language. Notably, the absence of training data does not necessarily result in poor intelligibility, suggesting the potential of such methods in addressing low-resource challenges. An emerging alternative approach leverages typological information as a non-parallel guidance mechanism for cross-linguistic transfer [20]. This method exploits systematic linguistic connections observed across various phonetic and structural levels, despite the significant diversity among languages. The present study explores a similar methodology for TTS in Indian languages.

Many languages historically lacked an independent script and later adopted existing scripts (Ex: Rajasthani using Devanagari script, Kui using Odia script). This is similar to the African scenario, where some languages use the Latin script or its modified version [21]. In contrast to existing work on zero-shot synthesis, our focus is on languages that use a script from one language family, but their phonotactics[1] belong to a different language family (more details in Section 2). Consequently, direct script-based parsing without considering the language properties is often inadequate. The Indian subcontinent is considered an area of linguistic convergence or "sprachbund" [22, 23], where extended contact between languages has led to a convergence in phonological, grammatical, and other features. Motivated by this, we perform zero-shot synthesis with a 3-stage approach:

1. We expand the phone-based common label set (CLS) representation for Indian languages [24] to cover missing

---

[*]These authors contributed equally.

[1]Phonotactics is the sequence of phones allowed in a language.

sounds in the unseen languages.

2. We analyse the linguistic and phonotactic properties of the unseen languages and apply the appropriate grapheme-to-phoneme rules.

3. A suitable monolingual TTS system is selected for zero-shot synthesis.

Zero-shot synthesis is performed on four languages, collectively spoken by more than 18 million individuals– Sanskrit (classical language), two dialects of Konkani, Maithili and Kurukh (endangered language). The first three are official languages of India. Evaluations demonstrate the efficacy of the proposed zero-shot synthesis and highlight the importance of a language-centric approach, particularly across language families.

The rest of the paper is organised as follows. Section 2 highlights the motivation of our proposed work. Section 3 provides a brief overview of the individual properties of the languages under consideration. Sections 4 and 5 present the proposed methodology and discuss the experiments carried out, respectively.

## 2. Motivation

Some languages adopt scripts from different language families while preserving distinct phonotactic characteristics. For example, Kui/Kuvi is written in the Odia script (IA) but belongs to the South-Central Dravidian group. Bodo is a Sino-Tibetan language and is written in the Devanagari script (used by IA languages such as Hindi and Marathi). As mentioned earlier, India is considered an area of linguistic convergence. Similarities in prosodic patterns, such as stress and intonation, enhance the perceived resemblance among Indian languages. The shared phonetic and syntactic patterns across Indian language families, particularly between IA and DR languages due to their Sanskrit roots, geographical proximity, and centuries of cultural exchange [25, 26], offer opportunities for cross-lingual transfer. This linguistic overlap, further reinforced by the influence of hegemonic languages like Hindi and English, may enable speech generation for low-resource languages using shared phonetic representation and text parsing tools. These capabilities support important initiatives like vernacular lecture translation [27] and BhashaVerse's 36-language translation framework [28], underscoring the importance of developing adaptable TTS systems for low-resource Indian languages.

## 3. Properties of Sanskrit, Konkani, Maithili and Kurukh

Sanskrit, a classical language, has significantly influenced Indian languages. Written in the Devanagari script and despite belonging to the IA language family, Sanskrit does not have *schwa* deletion[2]. Telugu, originating from Proto-Dravidian, has extensive Sanskrit influence, with similar vocabulary and grammatical structures [29]. Similarly, Kannada (of DR origin) also reflects Sanskrit influence [30]. Leveraging these properties, we hypothesise that DR language TTS models may better synthesise Sanskrit than IA models.

Konkani, an IA language spoken in the Konkan region (western coast of India), evolved from Old IA through Maharashtri Prakrit and Middle IA, retaining archaic features and a strong inflectional structure akin to Sanskrit. Its closest linguistic relative is Marathi, but high nasalisation makes it unique [31]. Konkani has various dialects and is written in multiple scripts, including Devanagari, Roman, Kannada, Malayalam, and Perso-Arabic. In this work, we have considered two dialects of Konkani– Maharashtrian (north coastal region) and Canara (south coastal region), written in Devanagari (IA) and Kannada (DR) scripts, respectively.

Maithili is an IA language primarily spoken in Bihar, Jharkhand, and Nepal[3]. It shares linguistic roots with Hindi but diverged centuries ago, forming a distinct identity despite lexical similarities [32]. Kurukh is a North DR language spoken in East India (mainly in Jharkhand, Chhattisgarh) by Kurukh (Oraon) and Kisan communities. As a DR language, Kurukh has distinct phonetic and grammatical features, but it has *schwa* deletion, which is characteristic of IA languages. Kurukh also has many phone borrowings, including phonemic glottal stop and complex verbal predicates from IA languages [33][4]. Both Maithili and Kurukh are written in Devanagari and share properties with IA languages [33, 34].

## 4. Proposed methodology

We now discuss the existing framework that supports zero-shot synthesis and present an overview of the proposed methodology.

### 4.1. Related framework for zero-shot synthesis

The number of phonemes in Indian languages is around 50. There are 15-18 vowels and 35-38 consonants [35]. The common label set (CLS) [24] provides a set of labels, wherein acoustically similar phones across 13 Indian languages are mapped together. The CLS, with a token size of 72, provides a very compact representation for training Indic multilingual systems. Although [15] proposed a unified phoneme inventory for South Asian languages, long and short vowels are mapped to a single unit ('इ', 'ई' are mapped to /i/ and 'उ', 'ऊ' to /u/), which is incorrect for Indian languages as meanings can change if used interchangeably.

The unified parser is a grapheme-to-phoneme (G2P) converter for 13 Indian languages, with distinct rules for IA and DR scripts [36]. *Schwa* deletion rules are applied for IA languages, unlike DR languages which are mostly written the way they are spoken. The unified parser converts text to CLS representation. Since the unified parser is generic to each language family, it fails for zero-shot synthesis of languages whose script belongs to one family, while the phonotactics belongs to another. This requires that the parser accounts for these linguistic anomalies and appropriate monolingual TTS systems are chosen for synthesis.

### 4.2. Updation of CLS for unseen languages

Coverage for all sounds (72) is ensured across various scripts by mapping missing and absent sounds to the closest equivalent[5]. For example, the diacritical *nukta/bindi*, which is present in scripts like Devanagari (used in lan-guages such as Kurukh, Hindi and Marathi), is not present in DR scripts. In such cases, the voiceless uvular plosive /q/ represented by 'क़' (*kq*) in Kurukh, Hindi, and Marathi undergoes substitution with the velar plosive /k/ through its base character 'क' in Devanagari,'ఙ' in Telugu and 'ಕ' in Kannada. Parallel neutralisation occurs for the voiced uvular fricative /ʁ/ denoted by 'ग़' (*gq*), which reduces to the voiced velar stop /g/, while the retroflex flap 'ड़' (*dxq*) becomes a dental flap /ɾ/ represented by '*dx*'. Other modifications and details on phonetic substitutions are provided in the referenced source. For consistency, the revised CLS will be henceforth referred to as CLS throughout this paper.

---

[2]*Schwa* deletion refers to the deletion of the inherent short vowel /ə/.

[3]http://lisindia.ciil.org/Maithili/Maithili.html
[4]http://www.language-archives.org/language/kru
[5]https://tts-synth.github.io/cls_map.html

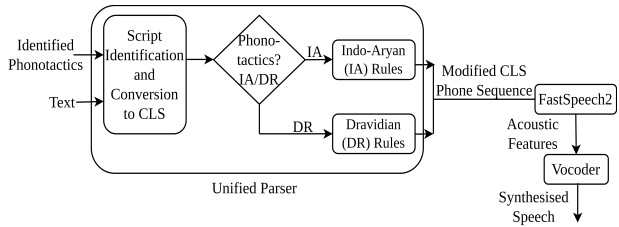

Figure 1: *Flowchart of the proposed approach: appropriate IA/DR rules are selected based on phonotactics of the target language*

### 4.3. Modifications to text parsing

A flowchart of the proposed approach is shown in Figure 1. Instead of directly applying the G2P rules based on the script, the text is first converted to a common representation (grapheme based) using CLS labels. Next, based on the phonotactics of the target/unseen language, the corresponding family-specific rules are applied. For example, since the phonotactics of Sanskrit is close to that of DR languages, DR rules (including *schw*a retention) are applied to the CLS obtained for the IA (Devanagari) text. Similarly, for Konkani dialects, IA rules are used for Maharashtrian Konkani, while DR rules are applied for Canara Konkani.

In Sanskrit, the visarga (○ः, /ḥ/) undergoes phonetic modification based on its position. At the end of a word, it is pronounced as a hard /h/ sound followed by a brief echo of the preceding vowel. If the preceding vowel is ऐ (/ai/) or औ (/au/), the echo corresponds to इ (/i/) or उ (/u/), respectively (Ex: कृष्णः (kṛṣṇaḥ) → kṛṣṇaha). When occurring before a sibilant (श, ष, स), the visarga is assimilated into the following consonant, effectively replacing it (Ex: अन्तःस्थ (antaḥstha) → antasstha).

For Maithili, a few parsing rules were added to account for phonetic change by way of epenthesis, i.e., backward transposition of final /i/ and /u/ in many words[3]. For example, अछि /əchi/ –> अइछ /əich/ ('is'), मधु /mədhu/ –> मउध /məudh/ ('honey'), बालु /ba:lu/ –> बाउल /ba:ul/ ('sand'). Also, unlike a DR language, Kurukh has borrowed deep aspirated /xʰ/ consonants (represented by *nukta*/dot below) like ख़ from IA languages [34], which gets taken care of by the parser. IA rules are applied to Maithili and Kurukh.

### 4.4. Selection of monolingual TTS system

The monolingual TTS system for synthesis is selected based on two criteria– (1) whether IA/DR rules were applied during parsing, (2) linguistic overlap based on geographical proximity, similarity in phone set, vocabulary and grammatical structures.

## 5. Evaluation of zero-shot speech synthesis

### 5.1. Datasets and synthesisers

Datasets for training monolingual TTS systems were sourced from [37][6]. Individual FastSpeech2 and HiFiGAN vocoders [3, 38, 39] were trained for Hindi (Hi), Kannada (Ka), Marathi (Ma) and Telugu (Te) using 10 hours of data each.

### 5.2. Experiments with zero-shot synthesis

Three systems were used for synthesising Sanskrit text– Hindi system with IA rules (Sys 1), Kannada (Sys 2) and Telugu (Sys 3) systems with DR rules. Maharashtrian Konkani was synthesised using a Marathi TTS system with IA rules (Sys 1) and a Kannada-based system with DR rules (Sys 2). Canara Konkani was synthesised using the same systems but exclusively with DR rules. Maithili, an IA language written in the Devanagari script, was synthesised using a Hindi TTS system with IA rules (Sys 1). Kurukh, a DR language also written in the Devanagari script, was synthesised using both Hindi (Sys 1) and Kannada (Sys 2) TTS systems, with IA rules applied. The performance of each system was evaluated through both subjective and objective measures.

### 5.3. Mean opinion score (MOS) Test for naturalness

Native speakers aged 18–50 evaluated the naturalness of synthesised speech on a 5-point Likert scale (5 being the best). In each test, 5 synthesised audio files per system (corresponding to out-of-domain text) and 5 ground-truth/recorded (GT) audio files were presented in random order. Hence, a total of 20 audio files were evaluated for Sanskrit, 15 each for Maharashtrian and Canara Konkani, 10 for Maithili and 15 for Kurukh. Studio-recorded files were used as GT in Sanskrit, Maharashtrian Konkani and Maithili. For Canara Konkani, the GT audio was from a native speaker's speech recorded in a quite environment (owing to the non-availability of studio-recorded data). For Kurukh GT, recorded Bible audio[7] was the only resource available. VoiceFixer [40] was used to remove feeble background music and noise.

Table 1: *MOS with zero-shot synthesis*

| Language (evaluator count) | GT | Sys 1 | Sys 2 | Sys 3 |
|---|---|---|---|---|
| Sanskrit (25) | 4.48 | 3.25 (Hi) | **4.12 (Ka)** | 3.67 (Te) |
| Maharashtrian Konkani (16) | 4.42 | **3.64 (Ma)** | 2.54 (Ka) | – |
| Canara Konkani (6) | 4.11 | 2.31 (Ma) | **3.34 (Ka)** | – |
| Maithili (24) | 4.56 | **3.72 (Hi)** | – | – |
| Kurukh (12) | 4.28 | **4.02 (Hi)** | 2.96 (Ka) | – |

Table 1 presents the results of the MOS test. It is clearly seen that the proposed approach produces more natural speech quality compared to systems that did not consider any external linguistic information. As expected, the MOS of Sanskrit synthesis using the Hindi system is lower than that with Telugu/Kannada systems. This is primarily due to *schwa* deletion with the Hindi synthesiser and IA rules, as illustrated in Figure 2. Aspiration is more prominent in Kannada compared to Telugu[8], which accounts for the comparatively lower score with the latter for Sanskrit.

---

[6] https://www.iitm.ac.in/donlab/indictts/database
[7] https://live.bible.is/bible/KRUDEV/MAT/1

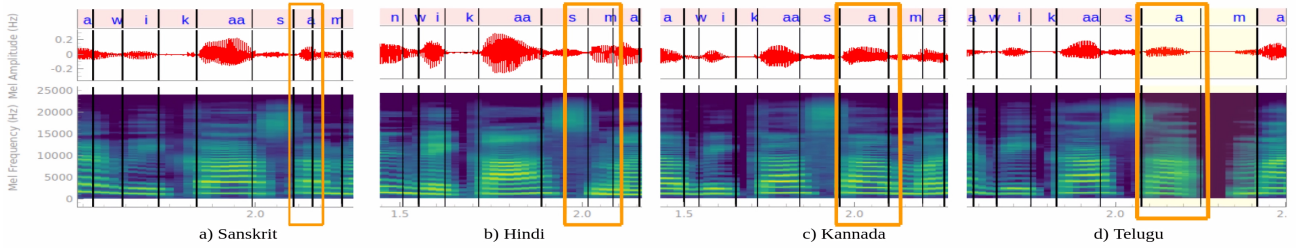

Figure 2: *Mel spectrogram of Sanskrit word '*विकास*' (w-i-k-aa-s-a) in different audio samples: a)Sanskrit GT, b)Hindi synthesised, c)Kannada synthesised, and d)Telugu synthesised. The schwa is preserved (highlighted) in the Kannada and Telugu audio, which is similar to the Sanskrit GT, while it is deleted in the Hindi audio.*

The Marathi system generated more natural speech than the Kannada system for Maharashtrian Konkani and vice versa for Canara Konkani. This highlights the significance of the proposed approach for dialect-based synthesis, particularly across language families. Kurukh, with the Hindi system, is more natural than its Kannada counterpart. However, Kurukh synthesis with both systems struggles with glottal stops. Moreover, question-type audio sounded more like an assertion for a few sentences, possibly due to unseen context. The MOS of the best system for Sanskrit (Sys 2) and Kurukh (Sys 1) are closer to that of GT, which is encouraging given that no data from that language was used for training/adaptation. The audio samples for all the experiments are available here[9].

### 5.4. SUS test for intelligibility

Table 2: *SUS test: WER with zero-shot synthesis*

| Language (evaluator count) | Sys 1 | Sys 2 | Sys 3 |
|---|---|---|---|
| Sanskrit (19) | 28.0% (Hi) | **4.6% (Ka)** | 15.0% (Te) |
| Maharashtrian Konkani(12) | **3.1% (Ma)** | 13.6% (Ka) | - |
| Canara Konkani(5) | 15.4% (Ma) | **5.6% (Ka)** | - |
| Maithili(20) | **5.4%(Hi)** | - | - |
| Kurukh(8) | **7.1% (Hi)** | 21.0% (Ka) | - |

Semantically unpredictable sentences (SUS) [41] (syntactically correct but semantically nonsensical) were synthesised and presented along with the text[10] for subjective evaluation. The sentences had about 8-15 words and followed subject-object-verb (SOV) word order. 10 sentences per system in each language were evaluated by native participants. Word error rate (WER) was calculated based on the wrongly pronounced words marked by evaluators.

---

[8]https://eprints.soas.ac.uk/28493/
[9]https://tts-synth.github.io/
[10]Providing the text to be marked ensured uniform bias due to few evaluators' poor technical literacy.

From Table 2, we see that the intelligibility follows a similar pattern to that of naturalness (Table 1). For Sanskrit synthesis, the WER is highest with the Hindi system (primarily due to *schwa* deletion), followed by the Telugu system (less prominent aspiration). Lower WER for Maharashtrian Konkani and Canara Konkani with Marathi and Kannada systems, respectively, is due to a larger vocabulary, linguistic and consonant cluster borrowings. The results indicate that the proposed framework generates reasonably intelligible output for the unseen languages.

### 5.5. Mel-cepstral distortion (MCD) scores

Table 3: *Mel-cepstral distortion (MCD) scores*

| Language | Sys1 | Sys 2 | Sys 3 |
|---|---|---|---|
| Sanskrit | 8.48 (Hi) | 7.15 (Ka) | 6.94 (Te) |
| Maharashtrian Konkani | 10.74 (Ma) | 8.28 (Ka) | - |
| Maithili | 7.90 (Hi) | - | - |
| Kurukh | 8.51 (Hi) | 8.28 (Ka) | - |

MCD score is an objective measure that determines the distortion in mel-cepstral features of a synthesised speech compared to a reference audio [42]. 30 audio files per system from the test set (not seen during training) were considered for evaluation. The MCD scores across different systems are presented in Table 3. For Sanskrit, the distortion observed with the DR family systems (Sys 2 and Sys 3) is lower compared to the Hindi system and the scores of Telugu and Kannada are similar. The difference in distortion between Kannada and Hindi systems for Kurukh is marginal. Canara Konkani was not considered, as the GT audio was noisy. Notably, the MCD score for Maharashtrian Konkani with the Marathi system is higher than that with the Kannada system. This can be attributed to artefacts introduced by the Marathi system, which needs further investigation.

## 6. Conclusions

In this work, we were able to generate intelligible and good-quality speech for unseen languages without any adaptation data. Linguistic overlap augmented with a modified shared CLS phone set and parsing rules has made zero-shot synthesis feasible even for languages with scripts and phonotactics from different language families. We also demonstrated the effectiveness of dialect-based zero-shot synthesis for dialects across language families with Konkani. This kind of language-centric approach can be extended to other low-resource languages and dialects, not confined solely to the Indian context.

# 7. References

[1] I. Office of the registrar general, *Paper 1 of 2018 Language, India, States And Union Territories, Table C-16*. Delhi: Census of India 2011, 2018. [Online]. Available: https://language.census.gov.in/eLanguageDivision_VirtualPath/eArchive/pdf/48.pdf

[2] A. van den Oord *et al.*, "Wavenet: A generative model for raw audio," in *Speech Synthesis Workshop*, 2016, p. 125.

[3] Y. Ren *et al.*, "FastSpeech 2: Fast and High-Quality End-to-End Text to Speech," in *ICLR*, 2021.

[4] Y. Chen *et al.*, "F5-TTS: A Fairytaler that Fakes Fluent and Faithful Speech with Flow Matching," 2024. [Online]. Available: https://arxiv.org/abs/2410.06885

[5] Y. A. Li *et al.*, "StyleTTS 2: Towards Human-Level Text-to-Speech through Style Diffusion and Adversarial Training with Large Speech Language Models," *NeurIPS*, vol. 36, pp. 19 594–19 621, 2023.

[6] J. Kim *et al.*, "Glow-TTS: A Generative Flow for Text-to-Speech via Monotonic Alignment Search," *NeurIPS*, vol. 33, pp. 8067–8077, 2020.

[7] V. Popov *et al.*, "Grad-TTS: A Diffusion Probabilistic Model for Text-to-Speech," vol. 139. PMLR, 2021, pp. 8599–8608.

[8] J. Kim *et al.*, "Conditional Variational Autoencoder with Adversarial Learning for End-to-End Text-to-Speech." PMLR, 2021, pp. 5530–5540.

[9] T. Saeki *et al.*, "Extending Multilingual Speech Synthesis to 100+ Languages without Transcribed Data," in *ICASSP*, 2024, pp. 11 546–11 550.

[10] F. Lux *et al.*, "Meta Learning Text-to-Speech Synthesis in over 7000 Languages," in *Interspeech*, 2024, pp. 4958–4962.

[11] E. Casanova *et al.*, "YourTTS: Towards Zero-Shot Multi-Speaker TTS and Zero-Shot Voice Conversion for everyone." PMLR, 2022, pp. 2709–2720.

[12] Z. Zhang *et al.*, "Speak Foreign Languages with Your Own Voice: Cross-Lingual Neural Codec Language Modeling," *arXiv preprint arXiv:2303.03926*, 2023.

[13] R. Joshi *et al.*, "Rapid Speaker Adaptation in Low Resource Text to Speech Systems using Synthetic Data and Transfer learning," in *PACLIC 37*. ACL, 2023, pp. 267–273.

[14] K. Azizah *et al.*, "Hierarchical Transfer Learning for Multilingual, Multi-Speaker, and Style Transfer DNN-Based TTS on Low-Resource Languages," *IEEE Access*, vol. 8, pp. 179 798–179 812, 2020.

[15] A. Gutkin, "Uniform Multilingual Multi-Speaker Acoustic Model for Statistical Parametric Speech Synthesis of Low-Resourced Languages." in *Interspeech*, 2017, pp. 2183–2187.

[16] O'Horan *et al.*, "Survey on the Use of Typological Information in Natural Language Processing." The COLING 2016 Organizing Committee, pp. 1297–1308.

[17] A. Prakash *et al.*, "Exploring the Role of Language Families for Building Indic Speech Synthesisers," *IEEE/ACM TASLP*, vol. 31, p. 734 – 747, 2023.

[18] ——, "Building Multilingual End-to-End Speech Synthesisers for Indian Languages," in *SSW*, 2019, pp. 194–199.

[19] A. Prakash and H. Murthy, "Generic Indic Text-to-Speech Synthesisers with Rapid Adaptation in an End-to-End Framework," in *SSW*, 2020, pp. 2962–2966.

[20] E. M. Bender, "On Achieving and Evaluating Language-Independence in NLP," *Linguistic Issues in Language Technology*, vol. 6, pp. 1–28, Oct. 2011.

[21] P. Ogayo, G. Neubig, and A. W. Black, "Building African Voices," in *Interspeech*, 2022, pp. 1263–1267.

[22] M. B. Emeneau, "India as a linguistic area," *Language*, vol. 32, no. 1, pp. 3–16, 1956. [Online]. Available: http://www.jstor.org/stable/410649

[23] N. Trubetzkoy, "Proposition 16," in *Actes du Premier Congrès International de Linguistes, a La Haye*. W. Sijthoff, 1930, 10–15 Apr 1928, pp. 17–18.

[24] B. Ramani *et al.*, "A common attribute based unified HTS framework for speech synthesis in Indian languages," in *SSW*, 2013, pp. 291–296.

[25] C. Masica, *The Indo-Aryan Languages*, ser. Cambridge Language Surveys. Cambridge University Press, 1993. [Online]. Available: https://books.google.co.in/books?id=J3RSHWePhXwC

[26] B. Krishnamurti, *The Dravidian Languages*, ser. Cambridge Language Surveys. Cambridge University Press, 2003. [Online]. Available: https://books.google.co.in/books?id=54fV7Lwu3fMC

[27] A. Prakash *et al.*, "Technology Pipeline for Large Scale Cross-Lingual Dubbing of Lecture Videos into Multiple Indian Languages," in *Interspeech*, 2023, pp. 3683–3684.

[28] V. Mujadia and D. M. Sharma, "BhashaVerse : Translation Ecosystem for Indian Subcontinent Languages," 2025. [Online]. Available: https://arxiv.org/abs/2412.04351

[29] T. K. Rao and T. V. Prased, "Comparative Analysis of Telugu and Sanskrit Languages," *KAAV International Journal of Science, Engineering Technology*, 2014.

[30] M. Mallikarjun, "Kannada versus Sanskrit: Hegemony, Power and Subjugation," *Language in India*, vol. 17, no. 8, 2017.

[31] S. Fadte *et al.*, "Empirical Analysis of Oral and Nasal Vowels of Konkani," 2023. [Online]. Available: https://arxiv.org/abs/2305.10122

[32] S. K. Jha, "Exploring the degree of similarities between Hindi and Maithili words from glottochronological perspective," *International Journal of Innovations in TESOL and Applied Linguistics*, vol. 5, 2019.

[33] M. Kobayashi and B. Tirkey, *The Kurux Language: Grammar, Texts, and Lexicon*, ser. Brill's Studies in South and Southwest Asian Languages, 2017, vol. 08, no. 8.

[34] A. Abbi, "Languages in Contact in Jharkhand. A Case of Language Conflation, Language Change and Language Convergence," in *Languages of tribal and indigenous peoples of India: the ethnic space*, 1997, no. 10, pp. 131—148.

[35] A. L. Thomas *et al.*, "Code-switching in Indic Speech Synthesisers," in *Interspeech*, 2018, pp. 1948–1952.

[36] A. Baby *et al.*, "A Unified Parser for Developing Indian Language Text to Speech Synthesizers," in *TSD*, 2016, pp. 514–521.

[37] ——, "Resources for Indian languages," in *TSD*, 2016, pp. 37–43.

[38] A. Prakash *et al.*, "Towards Developing State-of-The-Art TTS Synthesisers for 13 Indian Languages with Signal Processing Aided Alignments," in *ASRU*, 2023, pp. 1–8.

[39] J. Kong *et al.*, "HiFi-GAN: generative adversarial networks for efficient and high fidelity speech synthesis," in *NeurIPS*, 2020, pp. 17 022–17 033.

[40] H. Liu *et al.*, "VoiceFixer: A Unified Framework for High-Fidelity Speech Restoration," in *Interspeech*, 2022, pp. 4232–4236.

[41] C. Benoît *et al.*, "The SUS test: A method for the assessment of text-to-speech synthesis intelligibility using Semantically Unpredictable Sentences," *Speech Communication*, vol. 18, no. 4, pp. 381–392, 1996.

[42] R. Kubichek, "Mel-cepstral distance measure for objective speech quality assessment," in *IEEE Pacific Rim Conference on Communications Computers and Signal Processing*, vol. 1, 1993, pp. 125–128.